\documentclass[
]{ceurart}

\usepackage{listings}
\usepackage{algorithm}
\usepackage{algpseudocodex}
\usepackage{xcolor}  
\newcommand{\InlineComment}[1]{\quad$\triangleright$ \textit{\color{gray}#1}}

\begin{document}

\copyrightyear{2026}
\copyrightclause{Copyright for this paper by its authors.
  Use permitted under Creative Commons License Attribution 4.0
  International (CC BY 4.0).}

\conference{R. Campos, A. Jorge, A. Jatowt, S. Bhatia, M. Litvak (eds.): Proceedings of the Text2Story'26 Workshop, Delft (The Netherlands), 29-March-2026}

\title{Agenda-based Narrative Extraction: Steering Pathfinding Algorithms with Large Language Models}

\author[1]{Brian Felipe Keith-Norambuena}[%
orcid=0000-0001-5734-8962,
email=brian.keith@ucn.cl,
]
\cormark[1]
\author[2]{Carolina Inés Rojas-Córdova}[%
orcid=0000-0002-9751-3587,
]
\author[1]{Claudio Juvenal Meneses-Villegas}[%
orcid=0000-0003-1112-4925,
]
\author[3]{Elizabeth Johanna Lam-Esquenazi}[%
orcid=0000-0002-0388-4660,
]
\author[1]{Angélica María Flores-Bustos}[%
orcid=0009-0007-8989-6809,
]
\author[1]{Ignacio Alejandro Molina-Villablanca}[%
orcid=0009-0003-0535-5521,
]
\author[2]{Joshua Emanuel Leyton-Vallejos}[%
orcid=0009-0006-4535-623X,
]
\address[1]{Department of Computing and Systems Engineering, Universidad Católica del Norte, Antofagasta, Chile}
\address[2]{Department of Industrial Engineering, Universidad Católica del Norte, Antofagasta, Chile}
\address[3]{Department of Chemical and Environmental Engineering, Universidad Católica del Norte, Antofagasta, Chile}

\cortext[1]{Corresponding author.}

\begin{abstract}
  Existing narrative extraction methods face a trade-off between coherence, interactivity, and multi-storyline support. Narrative Maps supports rich interaction and generates multiple storylines as a byproduct of its coverage constraints, though this comes at the cost of individual path coherence. Narrative Trails achieves high coherence through maximum capacity path optimization but provides no mechanism for user guidance or multiple perspectives. We introduce agenda-based narrative extraction, a method that bridges this gap by integrating large language models into the Narrative Trails pathfinding process to steer storyline construction toward user-specified perspectives. Our approach uses an LLM at each step to rank candidate documents based on their alignment with a given agenda while maintaining narrative coherence. Running the algorithm with different agendas yields different storylines through the same corpus. We evaluated our approach on a news article corpus using LLM judges with Claude Opus 4.5 and GPT 5.1, measuring both coherence and agenda alignment across 64 endpoint pairs and 6 agendas. LLM-driven steering achieves 9.9\% higher alignment than keyword matching on semantic agendas ($p=0.017$), with 13.3\% improvement on \textit{Regime Crackdown} specifically ($p=0.037$), while keyword matching remains competitive on agendas with literal keyword overlap. The coherence cost is minimal: LLM steering reduces coherence by only 2.2\% compared to the agenda-agnostic baseline. Counter-agendas that contradict the source material score uniformly low (2.2-2.5) across all methods, confirming that steering cannot fabricate unsupported narratives.
\end{abstract}


\maketitle

\section{Introduction}
\label{sec:introduction}
Narrative extraction from document collections has become an area of interest for applications ranging from news summarization \cite{sojitra2024timeline,yadav2025news} to intelligence analysis \cite{ranade2022computational}. By organizing documents into coherent storylines, these methods help users understand relationships and progressions within large datasets \cite{keith2023survey,keith2026ina}. Our work falls within the event-based narrative extraction paradigm \cite{keith2023survey}, where we operationalize events at the document level: each document represents a single event, and a narrative is a temporally ordered, thematically coherent sequence of such document-events.

Prior work has proposed several approaches for extracting document-level storylines. Connect the Dots \cite{10.1145/1835804.1835884} finds chains between fixed endpoints that maximize word-based coherence. Narrative Maps \cite{keith2020narrative} extends this to directed acyclic graphs with multiple interconnected storylines, using linear programming to balance coherence with corpus coverage; its visualization interface supports rich user interaction through semantic operations like grouping events and adjusting connections. Other methods include Metro Maps \cite{shahaf2012metro}, which constructs multi-storyline visualizations, and newsLens \cite{laban2017newslens}, which builds parallel timelines across datasets. 

More recently, Narrative Trails \cite{german2025narrativetrails} operates on a coherence graph---where documents are nodes and edge weights quantify how well one document follows another---and finds the maximum capacity path, the route where the weakest connection is as strong as possible. This bottleneck formulation ensures no narrative transition falls below a quality threshold, producing highly coherent individual storylines but offering no mechanism for user guidance or alternative perspectives.

These approaches occupy different points in a design space defined by coherence, interactivity, and multi-storyline support. Narrative Maps enables interaction and naturally produces multiple storylines through coverage constraints, but its optimization can yield less coherent individual paths. Narrative Trails achieves high coherence through direct path optimization, but produces a single deterministic output with no way to incorporate user perspective or generate alternative framings of the same events.

In this paper, we introduce \textbf{agenda-based narrative extraction}, a method that bridges this gap. Our approach preserves the coherence properties of Narrative Trails while introducing interactivity through user-defined \textit{agendas}---natural language descriptions of the perspective the user wishes to explore. By integrating a large language model into the pathfinding process \cite{yao2023tree,herr2025llm}, we enable the algorithm to steer toward documents that support the specified agenda. Running the algorithm with different agendas yields different storylines through the same corpus, enabling the kind of multi-perspective exploration that Narrative Maps achieves through coverage constraints, but via user-specified framing rather than algorithmic diversification. Thus, our work addresses two research questions. First, we investigate whether LLM-guided steering can produce narratives that align with specified agendas while maintaining coherence (RQ1). Second, we examine the trade-off between agenda alignment and narrative coherence (RQ2).

The remainder of this paper is organized as follows. Section \ref{sec:related-work} reviews related work on narrative extraction and LLM-guided systems. Section \ref{sec:methods} describes our agenda-driven pathfinding method and evaluation framework. Section \ref{sec:results} reports our findings, including sensitivity analysis. Section \ref{sec:discussion} discusses implications and limitations, and Section \ref{sec:conclusion} concludes.

\section{Related Work}
\label{sec:related-work}
\textbf{Computational Narrative Extraction.}
Computational narrative extraction encompasses methods for identifying and structuring storylines from document collections \cite{keith2023survey,ranade2022computational,santana2023survey,keith2026ina}. Connect the Dots \cite{10.1145/1835804.1835884} formulates narrative extraction as finding chains between fixed endpoints that maximize word-based coherence constraints. Narrative Maps \cite{keith2020narrative} extends this to directed acyclic graphs with multiple interconnected storylines, using linear programming to balance coherence and coverage. Metro Maps \cite{10.1145/2339530.2339706} similarly constructs multi-storyline visualizations. The newsLens algorithm \cite{laban-hearst-2017-newslens} builds parallel timelines across entire datasets. Recent work has explored event-centric knowledge graphs for narrative construction \cite{yan2023narrative} and multilingual narrative extraction from news \cite{stefanovitch2025multilingual}. None of these methods, however, allow the user to inject a particular viewpoint or framing into the extraction itself.

\textbf{LLM-Guided Systems.}
LLMs have recently shown promise as ranking agents in retrieval pipelines \cite{zhu2025large,lewis2020retrieval}. Sun et al. \cite{sun2023llmreranking} found that GPT-class models re-rank search results competitively with supervised baselines; Qin et al. \cite{qin2023pairwise} pushed this further by showing that framing the task as pairwise comparisons yields better rankings than pointwise scoring. The LLM-as-a-judge paradigm has been validated for evaluating text quality \cite{zheng2023judging}, and Keith \cite{keith2025llm} specifically validates LLM judges as proxies for mathematical coherence measures in narrative extraction. We build on these findings but shift the goal: rather than improving relevance, we use the LLM to bias path selection toward a user-chosen narrative framing.

\textbf{Interactive Narrative Analytics.}
Interactive narrative analytics encompasses systems that allow analysts to explore and refine extracted storylines \cite{keith2026ina}; key challenges include bridging computational extraction methods with human sensemaking processes. The Narrative Maps Visualization Tool (NMVT) \cite{norambuena2025narrative} provides semantic interaction capabilities where analysts can group events, adjust connections, and iteratively refine narrative structures. Such systems support sensemaking workflows \cite{endert2012semantic} where analysts move between foraging for information and synthesizing coherent explanations \cite{pirolli2005sensemaking}. However, existing interactive systems require manual manipulation of graph structures; our approach provides interactivity through natural language agendas instead.

\textbf{Trade-offs in Narrative Extraction.}
Existing methods occupy different points in a design space defined by coherence quality, user interaction, and multi-storyline support \cite{keith2026ina}. Understanding these trade-offs motivates our approach. The Narrative Maps \cite{keith2020narrative} method formulates extraction as a linear programming problem optimizing coherence subject to coverage constraints. The 3MSI pipeline enables rich interaction (grouping events, adding/removing nodes), and coverage constraints naturally produce multiple storylines spanning different topic clusters. However, the LP formulation tends to produce less coherent individual paths than approaches that directly optimize path quality. The Narrative Trails \cite{german2025narrativetrails} method finds the maximum capacity path through a coherence graph, maximizing the minimum edge weight (the ``weakest link'') along extracted paths. This produces highly coherent storylines but offers no user interaction---a single deterministic output with no mechanism for incorporating user perspective or generating alternative framings. Our proposed approach bridges this gap: it preserves Narrative Trails' coherence optimization while enabling user guidance through natural language agendas. Different agendas applied to the same endpoints produce divergent paths through the corpus.

\section{Methodology}
\label{sec:methods}
\subsection{Background: Coherence Graphs and Maximum Capacity Paths}
Our method builds on the Narrative Trails framework~\cite{german2025narrativetrails} for storyline extraction. Given a corpus of documents $\mathcal{D}$, the algorithm constructs a coherence graph $G = (V, E, w)$ where nodes correspond to documents and edge weights represent pairwise coherence. Documents are first embedded using Sentence-BERT~\cite{reimers-2019-sentence-bert} into high-dimensional vectors. Since HDBSCAN~\cite{McInnes2017} is sensitive to the curse of dimensionality in high-dimensional spaces~\cite{aggarwal2001surprising}, embeddings are projected to 2D via UMAP~\cite{mcinnes2020umap} before clustering. While projecting to 2D loses information, it enables direct visualization of the embedding space, and the original Narrative Maps pipeline demonstrates that coherent narratives can still be extracted despite the low dimensionality \cite{keith2020narrative}. HDBSCAN is then applied to these 2D projections in soft clustering mode \cite{McInnes2017,campello2015hierarchical}, where each document receives a membership probability for every detected cluster rather than a single hard assignment. Each cluster is interpreted as a latent topic, so the result is a probability distribution over topics for each document. These topic membership distributions are then used as the topical component of the coherence function described below.

The coherence graph is constructed with temporal constraints that restrict edges to chronologically ordered pairs ($date(d_v) > date(d_u)$), ensuring the resulting graph is a directed acyclic graph. Following the original Narrative Trails algorithm \cite{german2025narrativetrails}, the graph is further sparsified by retaining only edges with coherence above a threshold derived from the maximum spanning tree.

A storyline from source $s$ to target $t$ is the path maximizing the minimum edge weight (the ``bottleneck''), solved via a modified Dijkstra's algorithm. This bottleneck formulation ensures that every transition in the extracted narrative meets a minimum quality threshold, unlike sum-of-weights optimization where a single weak link could break narrative flow while still producing a high total score. At each step, the algorithm pops the node with the highest capacity from a priority queue, identifies valid unvisited neighbors that can reach the target and would improve path capacity, and pushes the best neighbor to the queue. When the target node is popped, the path is returned.

\textbf{Coherence Function.} The coherence $\theta(d_u, d_v)$ between documents combines spatial proximity and topical similarity:
\begin{equation}
    \theta(d_u, d_v) = \sqrt{S(\hat{z}_u, \hat{z}_v) \cdot T(p_u, p_v)}
\end{equation}
where $S$ is angular similarity between 2D projections and $T = 1 - \text{JSD}(p_u, p_v)$ is topic similarity based on Jensen-Shannon divergence between cluster membership distributions \cite{keith2020narrative}.

\subsection{Agenda-Driven Pathfinding}
We extend the Narrative Trails algorithm to incorporate agenda-based steering. In the original algorithm described above, at each step, only the single highest-capacity neighbor is pushed to the priority queue. Our key modification is to instead consider the top-$k$ candidates by coherence and use an LLM to rank them based on agenda alignment, effectively replacing a greedy coherence-only selection with a multi-candidate LLM-guided selection.

\textbf{Algorithm Overview.} We show the pseudocode of the algorithm in Algorithm~\ref{alg:agenda-pathfinding}. Given a source document $d_s$, target document $d_t$, and agenda string $A$, the algorithm proceeds as follows. The search begins by initializing a priority queue with the source node using capacity $\infty$. At each step, the node with the highest priority is popped from the queue; if this node is the target, the path is returned. For the current node, all valid neighbors are identified: unvisited nodes that can reach the target and that would improve path capacity. The top-$k$ neighbors by edge coherence are selected for consideration. If $k > 1$ candidates remain, the LLM is queried to rank them by agenda alignment. Finally, all $k$ candidates are pushed to the queue prioritized by their LLM rank. 

\begin{algorithm}[!htb]
\caption{Agenda-Driven Narrative Extraction (Based on Narrative Trails)}
\label{alg:agenda-pathfinding}
\begin{algorithmic}[1]
\Require Coherence graph $G = (V, E, w)$; source $s$; target $t$; agenda $A$; candidate pool size $k$
\Ensure Path $P$ from $s$ to $t$, or $\varnothing$

\State $Q \leftarrow$ min-heap with entry $(0,\; s)$ \InlineComment{(priority, node)}
\State $\textit{cap}[s] \leftarrow \infty$;\; $\textit{parent}[s] \leftarrow \textsc{null}$;\; $\textit{processed} \leftarrow \emptyset$

\While{$Q \neq \emptyset$}
    \State $(\_,\; u) \leftarrow Q.\textsc{pop}()$
    \If{$u \in \textit{processed}$}
        \State \textbf{continue}
    \EndIf
    \State $\textit{processed} \leftarrow \textit{processed} \cup \{u\}$
    \If{$u = t$}
        \Return \textsc{ReconstructPath}($\textit{parent}$, $t$)
    \EndIf
    \State $\textit{visited} \leftarrow$ \textsc{ReconstructPath}($\textit{parent}$, $u$) \InlineComment{nodes on path to $u$}
    \State $C \leftarrow \emptyset$
    \For{each $(u, v) \in E$ with $v \notin \textit{visited}$}
        \If{\textsc{CanReachTarget}($G, v, t, \textit{visited}$)}
            \State $c_v \leftarrow \min\bigl(\textit{cap}[u],\; w(u,v)\bigr)$
            \If{$v \notin \textit{cap}$ \textbf{or} $c_v \geq \textit{cap}[v]$}
                \State $C \leftarrow C \cup \{(v,\; w(u,v),\; c_v)\}$
            \EndIf
        \EndIf
    \EndFor

    \State Sort $C$ by edge weight $w(u,v)$ descending \InlineComment{Top-$k$ selection}
    \State $C_k \leftarrow C[1 \mathrel{..} \min(k, |C|)]$

    \If{$|C_k| = 1$}
        \State $(v, \_, c_v) \leftarrow C_k[1]$
        \State $\textit{cap}[v] \leftarrow c_v$;\; $\textit{parent}[v] \leftarrow u$
        \State $Q.\textsc{push}\bigl((0,\; v)\bigr)$
    \ElsIf{$|C_k| > 1$}
        \State $R \leftarrow \textsc{LLMRank}(u,\; C_k,\; A,\; \textit{visited},\; t)$ \InlineComment{ordered list of nodes}
        \For{$i \leftarrow 1$ \textbf{to} $|R|$}
            \State $(v, \_, c_v) \leftarrow$ entry in $C_k$ for node $R[i]$
            \State $\textit{cap}[v] \leftarrow c_v$;\; $\textit{parent}[v] \leftarrow u$
            \State $Q.\textsc{push}\bigl((i - 1,\; v)\bigr)$ \InlineComment{lower value = higher priority; ties by insertion order}
        \EndFor
    \EndIf
\EndWhile
\Return $\varnothing$ \InlineComment{no path found}
\end{algorithmic}
\end{algorithm}

\textbf{LLM Ranking.} When multiple candidates are available, we construct a prompt that includes the agenda description $A$, the narrative context (titles of documents in the path so far), the current and target document titles, and the numbered candidate document titles. The LLM is instructed to return a JSON object with a ``ranking'' field containing the candidate numbers ordered from best to worst for advancing the agenda. We use Gemma 3 (12B) \cite{team2025gemma} via Ollama with temperature 0.3, top-p 0.9, and JSON output mode. After receiving the LLM ranking, candidates are pushed to the priority queue with priority determined directly by their LLM rank. The top-ranked candidate receives highest priority and is explored first, while lower-ranked candidates remain available for backtracking if the preferred path leads to a dead end or low-capacity bottleneck. The prompt is provided in Appendix~\ref{appendix:direct-ranking}.

\subsection{Baseline Methods}
We compare against two baseline methods that do not use LLM steering:

\textbf{Maximum Capacity.} The unmodified maximum capacity path algorithm from Narrative Trails \cite{german2025narrativetrails}, using $k=1$ and optimizing purely for coherence without any agenda consideration. This serves as the agenda-agnostic upper bound on coherence.

\textbf{Keyword Matching.} Uses TF-IDF similarity \cite{manning2008introduction} between candidate documents and the agenda text. At each step, candidates are scored by a weighted combination of coherence and agenda keyword overlap: $\text{score} = (1-\alpha) \cdot \text{coherence} + \alpha \cdot \text{tf-idf\_sim}$, with $\alpha = 0.5$. This baseline tests whether simple keyword overlap can achieve agenda alignment without semantic understanding.

\subsection{Evaluation Framework}
\textbf{Dataset.}
We evaluate on a corpus of 418 news articles covering the 2021 Cuban protests \cite{keith2023iui}, spanning December 2020 through late 2021. The articles cover the protests themselves, government response, Cuban-American diaspora reactions, U.S. policy responses, and international coverage. After applying date constraints and sparse coherence filtering with threshold $\tau = 1.0$, the coherence graph forms a connected directed acyclic graph suitable for pathfinding between temporally ordered endpoints.

\textbf{Endpoint Selection.}
We select source-target pairs that are temporally ordered and connected in the coherence graph, prioritizing pairs that are most distant in UMAP space. Sample sizes are determined by power analysis: 64 pairs for the main evaluation ($d=0.5$), with ablations using the 26, 21, or 17 most distant pairs ($d=0.8$, $d=0.9$, $d=1.0$ respectively).

\subsection{Agenda Definitions}
We define six agendas organized into three categories based on their relationship to the source material and the type of reasoning required to identify relevant articles.

\textbf{Simple Agendas.} These agendas contain keywords that appear literally in the corpus, making them amenable to keyword-based matching: \textit{Freedom Uprising} (``Cubans demanding freedom from communist rule'') and \textit{Diaspora Solidarity} (``Cuban-Americans rallying to support protesters in Cuba''). Terms like ``freedom,'' ``communist,'' and ``Cuban-Americans'' appear verbatim in article text.
\textbf{Semantic Agendas.} These agendas require inference beyond literal keyword matching \cite{turney2010frequency}: \textit{Regime Crackdown} (``Cuban regime violently suppressing protesters'') and \textit{Government Censorship} (``Cuban government controlling information through internet restrictions''). Terms like ``violently suppressing'' rarely appear verbatim; identifying relevant articles requires understanding that reports of arrests, beatings, or detention relate to this agenda.
\textbf{Counter Agendas.} These agendas contradict the source material and serve as negative controls: \textit{Protests Failing} (``Cuban protests losing momentum'') and \textit{Regime Popular} (``Cuban government maintaining popular support''). The corpus predominantly covers growing protests and regime opposition, so these agendas test whether methods can distinguish between achievable and impossible narrative goals.

\subsection{Multi-Judge Evaluation}
We evaluate extracted narratives using two frontier LLMs as judges: Claude Opus 4.5 (\texttt{claude-opus-4-5-20251101}) and GPT 5.1 (\texttt{gpt-5.1}). This dual-judge approach mitigates individual model biases and has been validated as an effective proxy for human evaluation of narrative coherence \cite{keith2025llm,zheng2023judging}. Each narrative is evaluated on two main dimensions: coherence and alignment.

In particular, for evaluation, each narrative is represented by concatenating the full text of all articles in the path, ordered according to the sequence discovered by the algorithm. This concatenated text is passed as the \texttt{\{narrative\_text\}} variable in the evaluation prompts (Appendices~\ref{appendix:coherence-eval} and \ref{appendix:alignment-eval}).

Coherence is assessed on a 1-10 scale across four sub-dimensions: logical flow, thematic consistency, temporal coherence, and narrative completeness. Alignment is assessed across five sub-dimensions: agenda support, persuasiveness, evidence strength, narrative direction, and bias effectiveness. The alignment prompt distinguishes surface-level topical relevance (scores 3-5) from true argumentative alignment (scores 7+). Complete prompts are in Appendix~\ref{appendix:coherence-eval} and \ref{appendix:alignment-eval}. For each metric, we report the mean of both judges' scores. Inter-rater reliability is strong for alignment (Pearson $r=0.88$, Spearman $\rho=0.93$) and moderate for coherence ($r=0.53$, $\rho=0.50$).

\section{Results}
\label{sec:results}
\subsection{Main Comparison: LLM-Driven vs. Baselines}
\textbf{Alignment by Agenda Type.} While overall alignment scores are similar across methods (Table~\ref{tab:combined-results}), method effectiveness varies by agenda type. On simple agendas where keywords appear literally in articles, keyword matching achieves the highest alignment, followed by the LLM-driven methods. On semantic agendas requiring inference, LLM-driven steering achieves the highest alignment, outperforming keyword matching. For the semantic agenda \textit{Regime Crackdown}, LLM-driven steering significantly outperforms keyword matching ($t=2.11$, $p=0.037$, Cohen's $d=0.37$).

\begin{table}[!htb]
\caption{Comparison of pathfinding methods across all agendas. Coherence and alignment scores are means across both judges (1-10 scale). Path length is the average number of articles in extracted narratives. Alignment by agenda type is shown for agenda-driven methods only.}
\label{tab:combined-results}
\centering
\begin{tabular}{lcccccc}
\hline
 & \multicolumn{3}{c}{Overall} & \multicolumn{3}{c}{Alignment by Agenda Type} \\
\cline{2-3} \cline{4-7}
Method & Coherence & Alignment & Path Length & Literal & Semantic & Counter \\
\hline
Maximum Capacity & 6.33 & 4.48 & 10.8 & --- & --- & --- \\
Keyword Matching & 6.28 & 4.64 & 11.8 & \textbf{6.91} & 4.78 & 2.24 \\
LLM-Driven ($k=5$) & 6.19 & 4.80 & 11.7 & 6.65 & \textbf{5.25} & 2.49 \\
\hline
\end{tabular}
\end{table}

\begin{figure}[!htb]
    \centering
    \includegraphics[width=0.9\columnwidth]{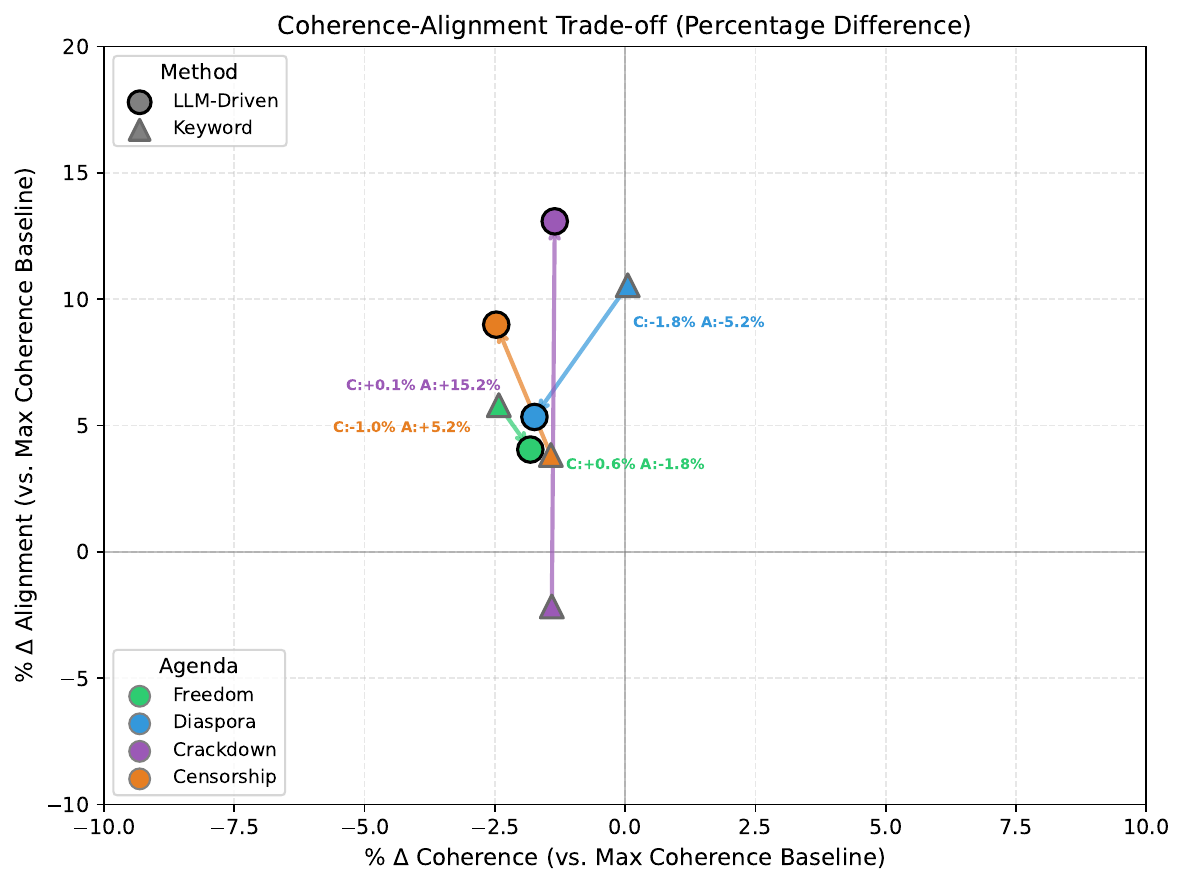}
    \caption{Coherence vs. alignment percentage differences with respect to the maximum capacity baseline for the keyword-based and LLM-based methods.}
    \label{fig:tradeoff}
\end{figure}

\textbf{Coherence Trade-off.} Figure~\ref{fig:tradeoff} shows the relationship between coherence and alignment across all 1,152 evaluations (64 endpoints $\times$ 6 agendas $\times$ 3 methods). LLM-driven steering reduces coherence by 0.14 points compared to the maximum capacity baseline (6.19 vs 6.33), a 2.2\% reduction. The correlation between coherence and alignment across all narratives is weak (Pearson $r\approx0.10$), suggesting that coherence and alignment are largely independent: narratives can achieve high alignment without sacrificing coherence, and vice versa.

\textbf{Counter-Agenda Validation.} All methods score uniformly low on counter-agendas (Table~\ref{tab:combined-results}, rightmost column). The range across methods is only 0.25 points (2.24-2.49), compared to spreads of 0.26 and 0.47 on simple and semantic agendas respectively. This confirms that steering methods cannot fabricate narratives contradicted by the source material. The \textit{Regime Popular} agenda scores lowest overall (mean 1.94), reflecting the corpus' predominant coverage of anti-government protests.

\subsection{Visualization of Steering Effects}
\begin{figure}[!htb]
    \centering
    \includegraphics[width=\columnwidth]{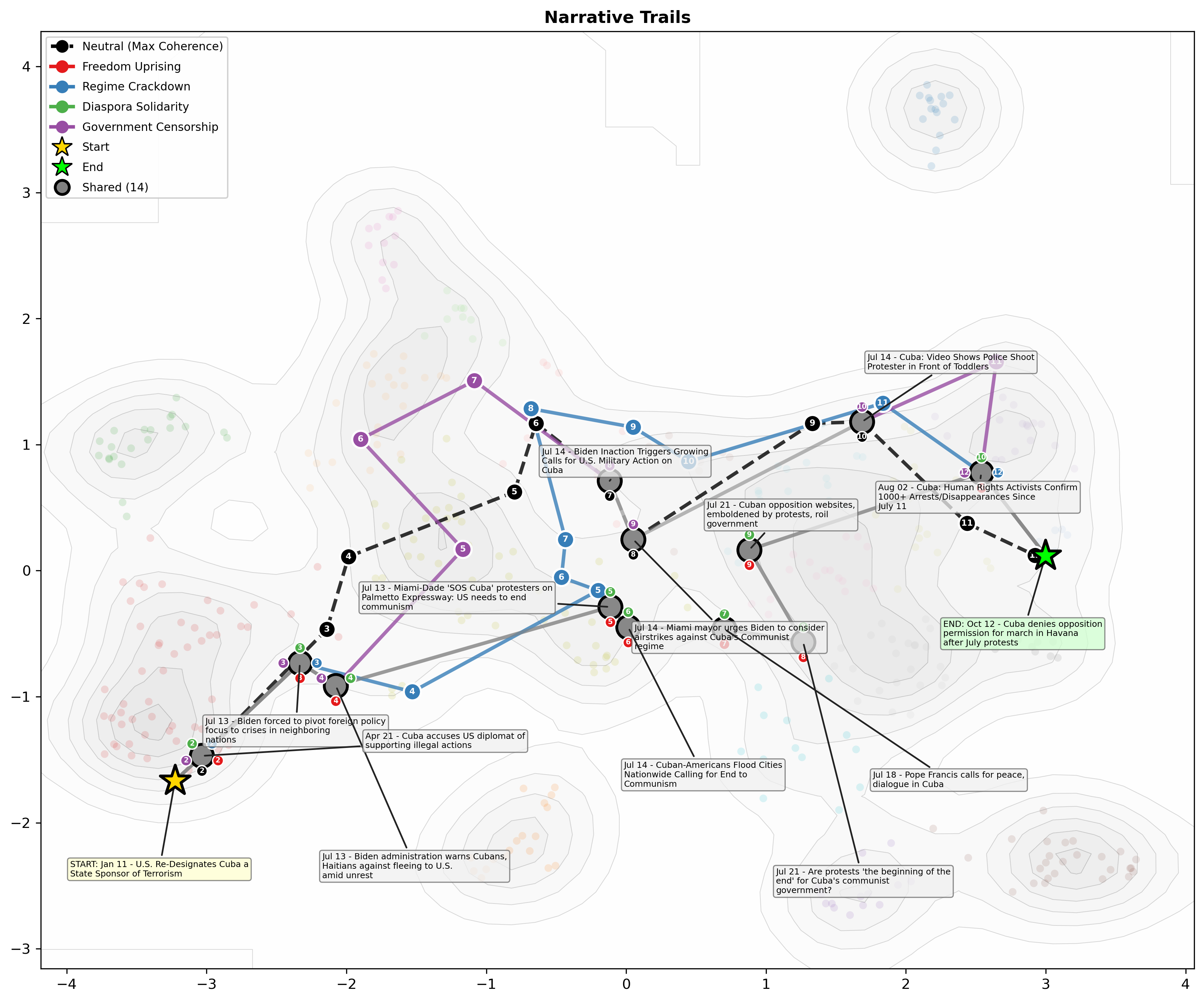}
    \caption{Narrative trails visualization showing how different agendas steer paths through the 2D embedding space from a trial run of the algorithm. The neutral baseline (optimal bottleneck coherence) is compared against four agenda-driven paths. Endpoints and shared nodes are highlighted. Background contours show article density in the UMAP projection space, illustrating that different agendas steer paths through different topical regions of the corpus---for instance, the Regime Crackdown path traverses denser clusters of government-response articles, while Freedom Uprising follows a trajectory through protest-coverage clusters.}
    \label{fig:narrative-trails}
\end{figure}

\begin{figure}[!htb]
    \centering
    \includegraphics[width=\columnwidth]{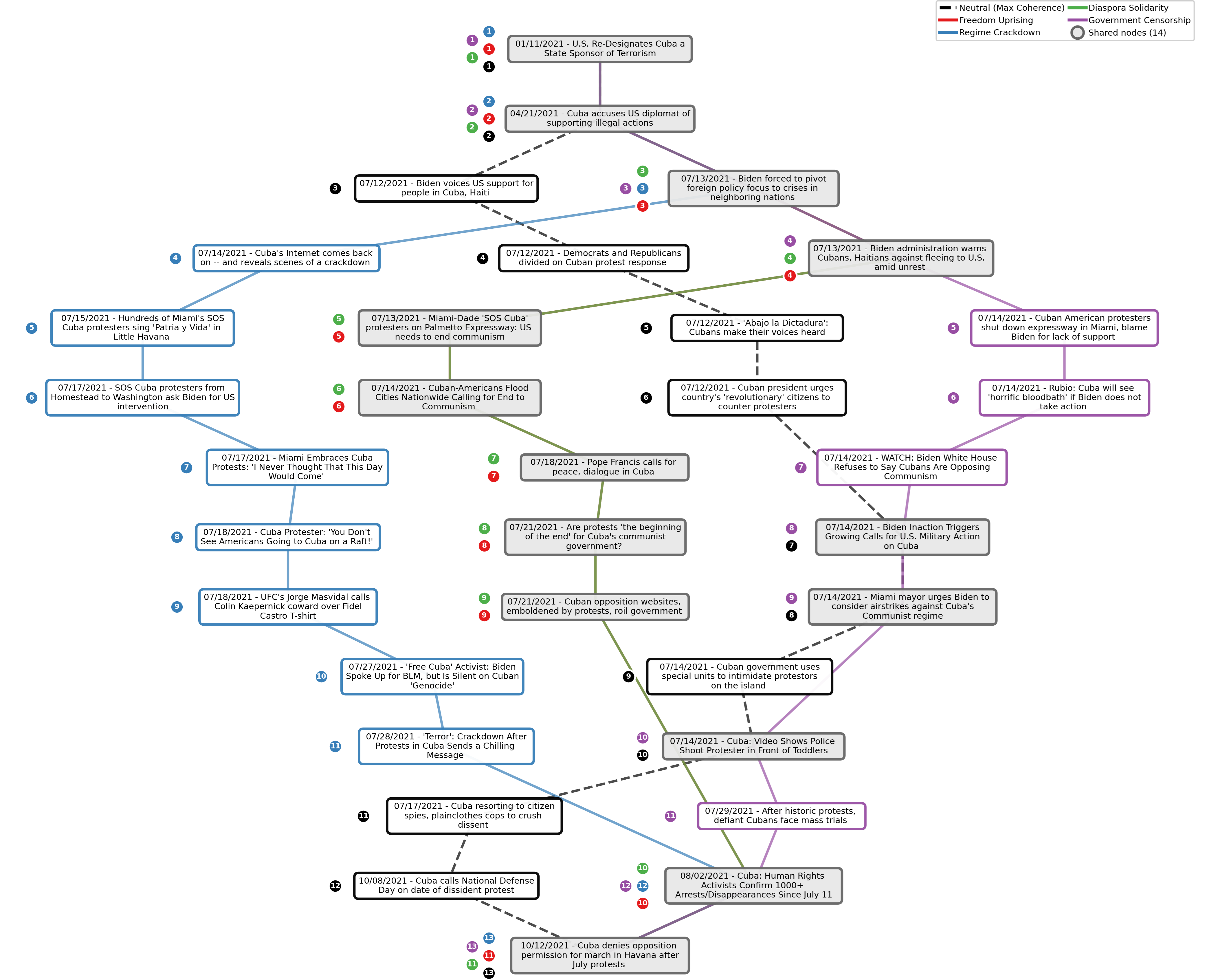}
    \caption{Equivalent narrative map showing the structure of extracted storylines from a trial run of the algorithm. All five paths (neutral baseline plus four agendas) flow from the source article at top to the target at bottom. Higher resolution version available here: \url{https://figshare.com/s/1aff7f3727b7ad4fa9dc}}
    \label{fig:narrative-map}
\end{figure}

Figures~\ref{fig:narrative-trails} and \ref{fig:narrative-map} visualize paths for a representative endpoint pair, showing how different agendas produce divergent trajectories through the corpus. Despite traversing different regions of the embedding space, all agenda-driven paths maintain coherence within 2.2\% of baseline. The Jaccard similarity between baseline and agenda-driven paths ranges from 0.25 to 0.50, confirming that steering produces substantially different document sequences. 

To illustrate, the Regime Crackdown path in Figures~\ref{fig:narrative-trails} and~\ref{fig:narrative-map} selects articles about arrests, police deployment, and mass trials---articles that do not contain the literal phrase ``regime crackdown'' but are semantically aligned with the agenda of violent suppression. Because the corpus consists predominantly of US news sources, the path also includes articles covering US political reactions to the crackdown; the LLM recognizes these as contextually relevant to the suppression narrative, even though they describe responses rather than the crackdown itself. This reflects a broader point: agenda-driven steering selects from the perspectives available in the data and cannot overcome source-level biases in the underlying corpus. Meanwhile, the Freedom Uprising path selects articles about protesters' demands and calls to end communist rule. Both paths are coherent sequences through the same corpus, but each emphasizes different facets of the underlying events, reflecting their respective agendas.

\subsection{Sensitivity Analysis}
We conducted an exploratory sensitivity analysis on candidate pool size ($k$=1-10), temperature (0.1-1.0), model size (270M-12B parameters). We also performed an ablation analysis on prompt design (varying specificity and including CoT). Due to smaller sample sizes, hyperparameter effects are not statistically significant but provide directional guidance. Coherence remained stable across all parameters. We used $k=5$, temperature 0.3, and Gemma 3 12B as defaults. Counterintuitively, vague agendas achieved higher alignment than specific ones, suggesting that overly detailed agendas may constrain the LLM's ability to recognize relevant articles.

\textbf{Chain-of-Thought Prompting.} The biggest effect we observed came from prompting style, not hyperparameters. Switching to CoT prompts \cite{wei2022chain} boosted alignment by 25.7\% (5.07 vs.\ 4.04, $p$=0.017) on a 17-endpoint subset without affecting coherence (both methods scored 6.15). However, CoT required 5.0$\times$ longer extraction time (161s vs 32s per narrative). The two prompting modes also chose different documents: Jaccard overlap between their paths averaged just 0.58, indicating that explicit reasoning leads to meaningfully different document selection. However, we note that the CoT prompt (Appendix~\ref{appendix:cot-ranking}) differs from the direct prompt (Appendix~\ref{appendix:direct-ranking}) in more than just the addition of step-by-step reasoning: it also includes an explicit definition of what it means for an article to ``fit'' a perspective and provides richer contextual framing. Therefore, the observed improvement cannot be attributed solely to chain-of-thought reasoning; the additional prompt structure could contribute to the effect.

\section{Discussion}
\label{sec:discussion}
\textbf{When LLM Steering Helps.} LLM-driven steering provides the largest benefit for semantic agendas where relevant articles cannot be identified through keyword matching. The 13.3\% improvement on \textit{Regime Crackdown} ($p=0.037$) demonstrates that LLMs can recognize when articles about arrests, detention, or police response relate to ``violent suppression'' even without those exact words \cite{brown2020language}. For simple agendas with literal keyword overlap, the computational overhead of LLM steering is not justified---keyword matching achieves comparable or better results at lower cost.

\textbf{Data Constraints.} The failure on counter-agendas across all methods demonstrates a fundamental constraint: no steering method can construct narratives unsupported by the source material. This is a feature, not a limitation. Agenda-driven steering amplifies perspectives present in the data; it does not fabricate them. The \textit{Regime Popular} agenda scores lowest because the corpus contains predominantly anti-government coverage---no amount of clever document selection can reverse this.

\textbf{Coherence Cost.} The 2.2\% coherence reduction from LLM steering is smaller than the natural variation across random endpoint pairs. In practical terms, a narrative scoring 6.19 on coherence is indistinguishable from one scoring 6.33. The weak correlation between coherence and alignment ($r=0.11$) further suggests that the trade-off is not fundamental: with appropriate candidate selection, narratives can be both coherent and aligned.

\textbf{Bridging Interactivity and Coherence.} Our results indicate that agenda-based steering can add user guidance to the Narrative Trails framework without substantially sacrificing its coherence properties, with only a 2.2\% reduction in coherence compared to the maximum capacity baseline on average. Meanwhile, the ability to generate different storylines by varying agendas provides multi-perspective exploration that baseline Narrative Trails lacks.

\textbf{Multiple Storylines via Agendas.} Different agendas produce different storylines through the same corpus. Running with ``Freedom Uprising'' versus ``Regime Crackdown'' agendas yields distinct narrative paths emphasizing different aspects of the Cuban protests. This capability resembles what Narrative Maps achieves through coverage constraints, but through user-specified perspective rather than algorithmic diversification. The sensitivity analysis shows that chain-of-thought prompting produces paths with only 0.58 Jaccard similarity to direct prompting paths for the same agenda, indicating that even the prompting strategy affects which documents are selected.

\textbf{Limitations.} Our evaluation uses a single corpus (2021 Cuban protests), and generalization to other domains requires further study~\cite{singhal2023domain}. We tested only six agendas organized into three categories, which does not represent the full agenda design space. The main evaluation uses 64 endpoint pairs, but ablation experiments use smaller subsets (17--26 pairs) determined by power analysis; while these provide directional guidance, they limit the statistical power of sensitivity and prompt-design comparisons. While inter-rater reliability is high for alignment ($r = 0.88$), LLM-based evaluation may have systematic biases not captured by agreement metrics~\cite{keith2025llm}. Human evaluation would provide complementary evidence and is planned for future work, particularly to validate that LLM judges' assessments of narrative quality align with analyst judgments in downstream sensemaking tasks. LLM steering adds latency compared to keyword matching, which would be detrimental for interactive applications, although this could be a limitation of the hardware used to perform the experiments.

\textbf{Ethical Considerations.} Agenda-driven narrative extraction could be misused to construct misleading narratives from selectively chosen documents \cite{wardle2017information}. Our work makes framing explicit and studiable rather than enabling manipulation \cite{friedman1996bias}. Understanding how narratives can be steered is a prerequisite for detecting and countering such steering in real-world information systems.

\section{Conclusion}
\label{sec:conclusion}
We introduced agenda-based narrative extraction, a method that bridges the gap between the interactivity of Narrative Maps and the coherence of Narrative Trails. By integrating LLMs into the maximum capacity pathfinding algorithm, our approach enables user-guided steering through natural language agendas while preserving the coherence properties of the underlying algorithm. The approach brings two capabilities to the Narrative Trails framework: interactivity through natural language agendas, allowing users to express their perspective without manipulating graph structures; and multiple storylines through agenda variation, enabling exploration of different framings of the same events.

Experiments on a news corpus demonstrate that LLM-driven steering achieves 9.9\% higher alignment than keyword matching on semantic agendas ($p=0.017$), with a coherence cost of only 2.2\%. Counter-agendas score uniformly low (2.2-2.5) regardless of method, confirming that steering amplifies perspectives present in the data rather than fabricating unsupported narratives. Sensitivity analysis reveals that performance is robust across hyperparameter variations, with chain-of-thought prompting offering alignment gains at increased computational cost.

\begin{acknowledgments}
This research is funded by the ANID FONDEF ID25I10072 Project "Narrative Panopticon: Intelligent Platform For Mapping And Monitoring Information Narratives From Multi-Source Data Streams." We also thank the \textit{coreDevX} team for supporting the Narrative Panopticon project. This work has also been supported by the ANID FONDECYT 11250039 Project "Interactive Narrative Analytics" and by Project 202311010033-VRIDT-UCN.
\end{acknowledgments}

\section*{Declaration on Generative AI}
During the preparation of this work, the authors used Claude to refine sections and support literature review activities. Additionally, Writefull integrated in Overleaf was used to improve writing quality and readability. After using these tools/services, the authors reviewed and edited the content as needed and take full responsibility for the content of the article.

\bibliography{sample-ceur}

\appendix

\section{Prompts}
\label{appendix:prompts}
This appendix contains the prompts used for LLM-driven pathfinding and evaluation.

\subsection{Direct Ranking Prompt}
\label{appendix:direct-ranking}
The following prompt is used to rank candidate documents by agenda alignment during pathfinding:

\begin{lstlisting}[basicstyle=\scriptsize\ttfamily,breaklines=true]
You are helping build a narrative that aligns with this agenda: {agenda}
Current narrative context:
{context_titles}
Current article: {current_article}
Target article: {target_article}
Rank ALL of the following articles from BEST to WORST based on how well they advance the agenda while helping reach the target article.
Options:
{numbered_candidates}
You MUST respond with a JSON object containing your ranking. Example format:
{"ranking": [3, 1, 2]}
The "ranking" field must be an array containing ALL numbers from 1 to {len(candidates)} in order from best to worst.
\end{lstlisting}

\subsection{Chain-of-Thought Ranking Prompt}
\label{appendix:cot-ranking}
The chain-of-thought variant adds explicit reasoning steps:

\begin{lstlisting}[basicstyle=\scriptsize\ttfamily,breaklines=true]
You are building a news narrative that supports a specific perspective.
THE PERSPECTIVE: "{agenda}"
An article "fits" if it provides evidence, examples, or context that reinforces this perspective. An article "doesn't fit" if it contradicts, undermines, or is irrelevant to this perspective.
STORY SO FAR:
{path_context}
CURRENT: {current_article}
DESTINATION: {target_article}
OPTIONS:
{numbered_candidates}
Think step by step before ranking:
STEP 1 - PERSPECTIVE ALIGNMENT: For each option, briefly explain whether it supports, contradicts, or is neutral to the perspective.
STEP 2 - PATH TO DESTINATION: Which options help us eventually reach the destination article?
STEP 3 - RANKING: Based on the above, rank ALL options from BEST to WORST for advancing the perspective while reaching the destination.
You MUST respond with a JSON object containing your reasoning and ranking. Example format:
{"reasoning": "your step-by-step analysis here", "ranking": [3, 1, 2]}
\end{lstlisting}

\subsection{Coherence Evaluation Prompt}
\label{appendix:coherence-eval}
The following prompt is used by LLM judges to evaluate narrative coherence:

\begin{lstlisting}[basicstyle=\scriptsize\ttfamily,breaklines=true]
You are evaluating the coherence of a sequence of news articles.
Rate the following narrative on these criteria (1-10 scale):
1. **Logical Flow**: How well do the articles connect logically?
2. **Thematic Consistency**: How consistent are the themes throughout?
3. **Temporal Coherence**: Do the events follow a reasonable timeline?
4. **Narrative Completeness**: Does the sequence tell a complete story?
Narrative sequence:
{narrative_text}
BE FAIR BUT CRITICAL. A score of 5 means "adequate", 7 means "good", 10 means "exceptional". Most narratives should score between 4-7.
Respond in JSON format:
{
    "logical_flow": <int 1-10>,
    "thematic_consistency": <int 1-10>,
    "temporal_coherence": <int 1-10>,
    "narrative_completeness": <int 1-10>,
    "overall_coherence": <float>,
    "explanation": "<brief explanation>"
}
\end{lstlisting}

\subsection{Agenda Alignment Evaluation Prompt}
\label{appendix:alignment-eval}
The following prompt is used by LLM judges to evaluate agenda alignment:

\begin{lstlisting}[basicstyle=\scriptsize\ttfamily,breaklines=true]
You are evaluating whether a sequence of news articles SUPPORTS and ARGUES FOR a specific agenda.
**CRITICAL DISTINCTION**: You must distinguish between:
1. **Surface-level relevance** (NOT alignment): Articles that share keywords/topics with the agenda but don't actually support it
2. **True argumentative alignment** (IS alignment): Articles that provide evidence, arguments, or framing that advances the agenda's claims
**IMPORTANT**: This narrative may or may not have been selected to support this agenda. Many narratives will NOT align with the agenda - be critical and look for specific evidence.
**Agenda**: {agenda}
**What constitutes TRUE agenda support (high scores):**
- Articles provide EVIDENCE or ARGUMENTS that support the agenda's claims
- The sequence builds a logical CASE for the agenda's conclusion
- Articles frame events in ways that FAVOR the agenda's viewpoint
- The cumulative effect PERSUADES toward the agenda's position
**What does NOT constitute support (low scores):**
- Articles merely MENTION the same topics, people, or keywords as the agenda
- Articles that are about the same subject but present NEUTRAL or BALANCED coverage
- Articles that CONTRADICT the agenda, even if they share vocabulary
- Generic news coverage that happens to use similar terms
Rate the narrative on these criteria (1-10):
1. **Agenda Support**: Do articles ARGUE FOR or provide EVIDENCE supporting the agenda's claims?
2. **Persuasiveness**: Does the sequence build a CONVINCING CASE?
3. **Evidence Strength**: How strong and RELEVANT is cumulative evidence FOR the agenda?
4. **Narrative Direction**: Does the sequence PROGRESS toward validating the agenda's conclusions?
5. **Bias Effectiveness**: Does the selection create FAVORABLE framing for the agenda?
Narrative sequence:
{narrative_text}
BE CRITICAL. Ask yourself: "Would reading these articles convince someone of the agenda's position?" Most topically-related narratives should score 3-5. Only narratives with genuine argumentative support should score 7+.
Respond in JSON format:
{
    "agenda_support": <int 1-10>,
    "persuasiveness": <int 1-10>,
    "evidence_strength": <int 1-10>,
    "narrative_direction": <int 1-10>,
    "bias_effectiveness": <int 1-10>,
    "overall_alignment": <float>,
    "explanation": "<brief explanation>"
}
\end{lstlisting}

\end{document}